\documentclass[a4paper,fleqn]{cas-dc}

\usepackage[numbers]{natbib}
\usepackage{todonotes}
\usepackage{mathtools}
\usepackage{url}
\usepackage{siunitx}
\usepackage{amsmath,amssymb,relsize}
\usepackage{todonotes}
\usepackage{multicol}
\usepackage{caption}
\usepackage{enumitem}
\usepackage[ruled,vlined,linesnumbered]{algorithm2e}

\newcommand{\abs}[1]{|#1|}

\begin{document}
\let\WriteBookmarks\relax
\def\floatpagepagefraction{1}
\def\textpagefraction{.001}
\shorttitle{Learning from Experience for Rapid Generation of Local Car Maneuvers}
\shortauthors{P. Kicki et~al.}

\title [mode = title]{Learning from Experience for Rapid Generation of Local Car Maneuvers}                      
\tnotemark[1]

\tnotetext[1]{This work is partially funded under the project "Advanced driver assistance system (ADAS) for precision maneuvers with single-body and articulated urban buses", co-financed from the European Union from the European Regional Development Fund within the Smart Growth Operational Programme 2014-2020 (contract No. POIR.04.01.02-00-0081/17-01).}


\author[1]{Piotr Kicki}[type=editor,
                        auid=000,bioid=1,
                        orcid=0000-0003-0097-6612]
\cormark[1]
\ead{piotr.kicki@put.poznan.pl}

\credit{Concept, Software Development, Experiments, Writing}

\address[1]{Institute of Robotics and Machine Intelligence, Pozna\'n University of Technology, ul. Piotrowo 3A, 60-965, Pozna\'n, Poland}

\author[1]{Tomasz Gawron}[orcid=0000-0003-0975-6077]
\ead{tomasz.gawron@put.poznan.pl}

\credit{Data generation, Control algorithm, Fruitful discussions}

\author[1]{Krzysztof Ćwian}
\ead{krzysztof.cwian@put.poznan.pl}

\credit{Data acquisition and preprocessing, Real-world maps generation}

\author[]{Mete Ozay}
\ead{meteozay@gmail.com}

\credit{Writing, Methodology, Fruitful discussions}

\author[1]{Piotr Skrzypczyński}[type=editor,
                        auid=000,bioid=4,
                        orcid=0000-0002-9843-2404]
\ead{piotr.skrzypczynski@put.poznan.pl}

\credit{Writing, Concept of the experiments, Methodology, Fruitful discussions}

\cortext[cor1]{Corresponding author}

\begin{abstract}
Being able to rapidly respond to the changing scenes and traffic situations by generating feasible local paths is of pivotal importance for car autonomy.
We propose to train a deep neural network (DNN) to plan feasible and nearly-optimal paths for kinematically constrained vehicles in small constant time. Our DNN model is trained using a novel weakly supervised approach and a gradient-based policy search.
On real and simulated scenes, and a large set of local planning problems, we demonstrate that
our approach outperforms the existing planners with respect to the number of successfully completed tasks. While the path generation time is about 40 ms, the generated paths are smooth and comparable to those obtained from conventional path planners.
\end{abstract}




\maketitle

\section{Introduction}

Self-driving cars utilized in city scenarios often require to quickly plan
local maneuvers taking into consideration the constraints imposed
by the kinematics of cars and geometry of scene perceived by vehicle's sensors.
Although state-of-the-art path planning methods are considered sufficient for
autonomous vehicles \cite{motionreview}, their ability to handle highly constrained
planning cases comes at a computational cost that is too high for rapid maneuver planning.
In contrast, parking or overtaking maneuvers are relatively easy for
experienced human drivers, who intuitively generate
nearly optimal paths of their vehicles in a short time horizon, exploiting the prior experience on how to perform these maneuvers \cite{groeger}.
Unfortunately, conventional path planners \cite{lavallebook} do not integrate the prior experience
that comes from learning on similar cases solved before.


\begin{figure*}[th]
    \centering
    \includegraphics[width=0.95\linewidth]{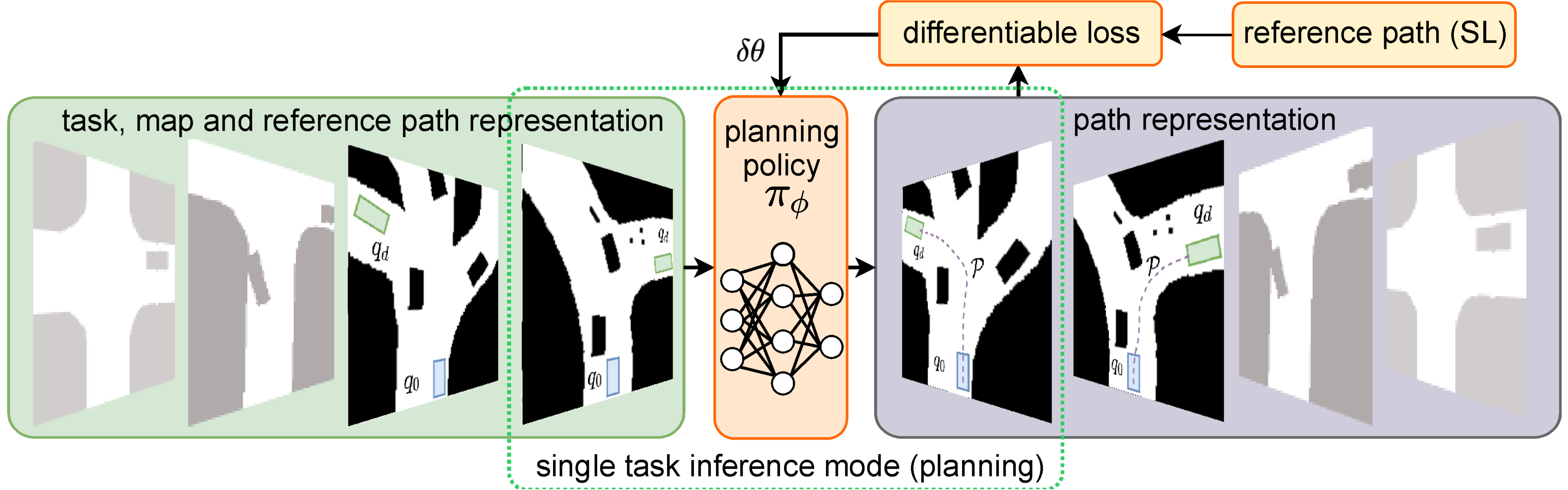}
    \caption{A conceptual scheme of the rapid path generation system. We use gradient-based policy search for training the proposed neural network to plan feasible paths.}
    \label{fig:scheme}
\end{figure*}

Therefore, in this paper, we investigate how to learn a path planning policy,
which is then used to rapidly generate feasible and nearly-optimal local
paths for typical maneuvers of a car-like vehicle, being subject to severe kinematic constraints.
For this purpose, we propose a deep neural network, which generates a feasible path by approximating the oracle planning function considering the representation of the task (Figure~\ref{fig:scheme}).
Our approach is not intended to be a standalone planner, as it is not complete.
It should be invoked by a higher-level planner to speed up operation in constrained scenarios similar to those already seen.
The proposed neural network model is trained using a gradient-based policy search approach \cite{policysurvey} in a reinforcement learning framework with a differentiable loss function.
Although we use reference paths in the evaluation of the cost function, these paths are generated automatically by a complete path planning algorithm and are used only to retreat from the collision areas.
Hence they do not constitute a performance upper bound for the proposed method, which can find paths that are shorter and/or smoother than the reference ones.
In principle, any planning algorithm able to handle the constraints present in our problem can be applied to generate the reference paths. In the experiments demonstrated in this article, we have chosen a variation of the State Lattices method \cite{SL} due to its practical advantages, such as reliability and fast generation of feasible plans.
The main contribution of this paper is twofold:
\begin{enumerate}
    \item  An approach for rapid path generation under differential constraints by approximating the oracle planning function using a novel neural network.
    \item A novel differentiable loss function which penalizes infeasible paths,
since they violate constraints imposed either by the vehicle kinematics or environment maps.
\end{enumerate}

Moreover, we propose a dataset of urban environment local maps (based on real sensory data)
and motion planning scenarios that can be used for training and evaluation of local planners for self-driving cars\footnote{Both code and the dataset are available at \url{https://github.com/pkicki/neural_path_planning}}.
Using this dataset, we demonstrate the generalization abilities of the learned model of our network, illustrate its advantages over selected state-of-the-art planning algorithms, and quantitatively analyze the accuracy of the proposed solution, planning times, and properties of the generated paths.

\section{Related Work}
Early works on self-driving cars used search-based algorithms, such as the A$^*$, in a quantized space \cite{thrun}.
As path planning for car-like vehicles involves high dimensional search spaces,
sampling-based algorithms are considered, due to their computational efficiency \cite{karaman}.
The basic Rapidly exploring Random Trees (RRT) algorithm provides no guarantee of the path quality, but RRT$^*$ \cite{karaman} extends this algorithm to an asymptotically optimal version, while Informed RRT$^*$ \cite{informed} incorporates a heuristic into the RRT$^*$ algorithm to improve the convergence rate.
The more recent Stable Sparse Random Trees \cite{SST} uses a much more sparse tree, still being asymptotically optimal. 
Batch Informed Trees (BIT$^*$) \cite{bit2015} is an asymptotically optimal anytime algorithm,
which quickly finds feasible solutions, and if the computational time permits, then converges towards the optimal one.
Its latest modification, Advanced BIT$^*$ \cite{abit2020}, applies advanced search techniques to balance between exploration and exploitation of the sampled state space.
An alternative solution is the Adaptively Informed Trees (AIT$^*$) algorithm \cite{ait2020}.
AIT$^*$ uses asymmetric bidirectional search to exploit a problem-specific heuristic,
which is updated during the search.
A~number of the recent sampling-based algorithms are implemented in the Open Motion Planning Library (OMPL) \cite{ompl}.
Unfortunately, if nonholonomic constraints and steering angle limits are present in the vehicle kinematics, then the performance of sampling-based methods degrades \cite{NonholonomicRRT}.
In contrast, the State Lattices algorithm \cite{SL} is relatively fast while planning paths for vehicles with nonholonomic constraints. However, if Dubins curves are applied as motion primitives, then the vehicles have to stop to make turns, whereas using smoother primitives results in a much higher computational cost.

Planning frameworks that use optimization and control methods have been proposed to cope with moving objects and motion prediction.
A planner for aggressive car maneuvers based on a version of RRT$^*$
and the non-linear Model Predictive Control method was presented in \cite{arab},
while a Constrained Iterative Linear Quadratic Regulator was used in \cite{ilqr}
to solve a planning problem with moving obstacles for on-road driving.
Unfortunately, these methods are computationally intensive and expensive for development and maintenance, because they require tedious manual parameter tuning for different scenarios.

The time-consuming manual tuning process implemented in the search and optimization-based path or trajectory planning methods is a consequence of the fact that none of these methods can exploit the similarity between the task at hand and the tasks solved before.
However, there are works that leverage previous experience in motion planning for manipulation.
An Experience Graph which represents the underlying connectivity of the space is
constructed from the previous planning episodes in \cite{egraphs} to accelerate motion planning whenever possible.
Demonstrated trajectories are used in \cite{experience} to quickly predict feasible trajectories
for novel scenarios in manipulation.
This approach applies a situation descriptor to find similar tasks and descriptions of the trajectories in the task space.
Similarly, entire paths are stored, and then are re-used or repaired the existing ones to reduce the planning time in the manipulation in \cite{berenson}.

The success of deep learning in robotics \cite{dlimits} increased the interest in
path and motion planning methods entirely learned from data.
Driving data collected from demonstrations of a human expert can be used to
learn a driving policy in urban scenarios with deep learning, as shown in \cite{imitation}.
A~convolutional neural network model is trained to predict ego vehicle's future trajectory
points taking as input a bird-view image obtained from the CARLA simulator.
In real-world scenarios, a perception module that transforms raw sensory data into
the symbolic bird-view image has to be deployed on the input of this system,
thus the final performance depends on this module to much extent.
A jointly learnable behavior and trajectory planner for self-driving vehicles was proposed in \cite{urtasun}.
This planner learns a shared cost function used to select behaviors that handle traffic rules
and generate vehicle trajectories employing a continuous optimization solver.
Nevertheless, it learns from a database of human driving recordings rather than from its own experience.

The deep learning paradigm has been used in the context of learning from
previous experience in manipulation tasks.
The recent paper \cite{next} tackles the problem of learning from previous experience to
boost path planning in high-dimensional continuous state and action spaces.
It extends a tree-based sampling algorithm by introducing learnable neural components that
bias the sampling towards more promising regions.
Recurrent neural networks were employed in the OracleNet \cite{oracle2019} to
generate new paths using paths obtained from demonstrations.
While this approach was an inspiration for our work, it differs significantly
from the planner presented in this paper, because it does not use any environment model,
relying on the demonstrated paths.
Such an approach is rather infeasible for vehicles in urban scenarios, where the
state space may differ considerably for particular maneuvers and local scenes.
Therefore, we use the reinforcement learning schema instead of direct demonstrations.
The OracleNet concept was then extended to the Motion Planning Networks (MPNet) framework \cite{mpnet},
which generates collision-free paths for the given start and goal states (configurations)
directly taking point clouds as input.
MPNet generalizes to unseen environments, but it still learns through imitation.
The StrOLL algorithm \cite{stroll} that tackles the problem of lazy search in graphs for path planning
also uses the notion of ``oracle'', but it learns planning policies by imitating an oracle that
has full knowledge about the environment map at training time, and can compute optimal decisions.
This algorithm does not consider non-holonomic constraints or path smoothness,
thus it is not directly applicable to our problem.

\section{Problem Definition}
\label{sec:problem_formulation}

\begin{figure*}[th]
    \centering
    \includegraphics[width=0.8\linewidth]{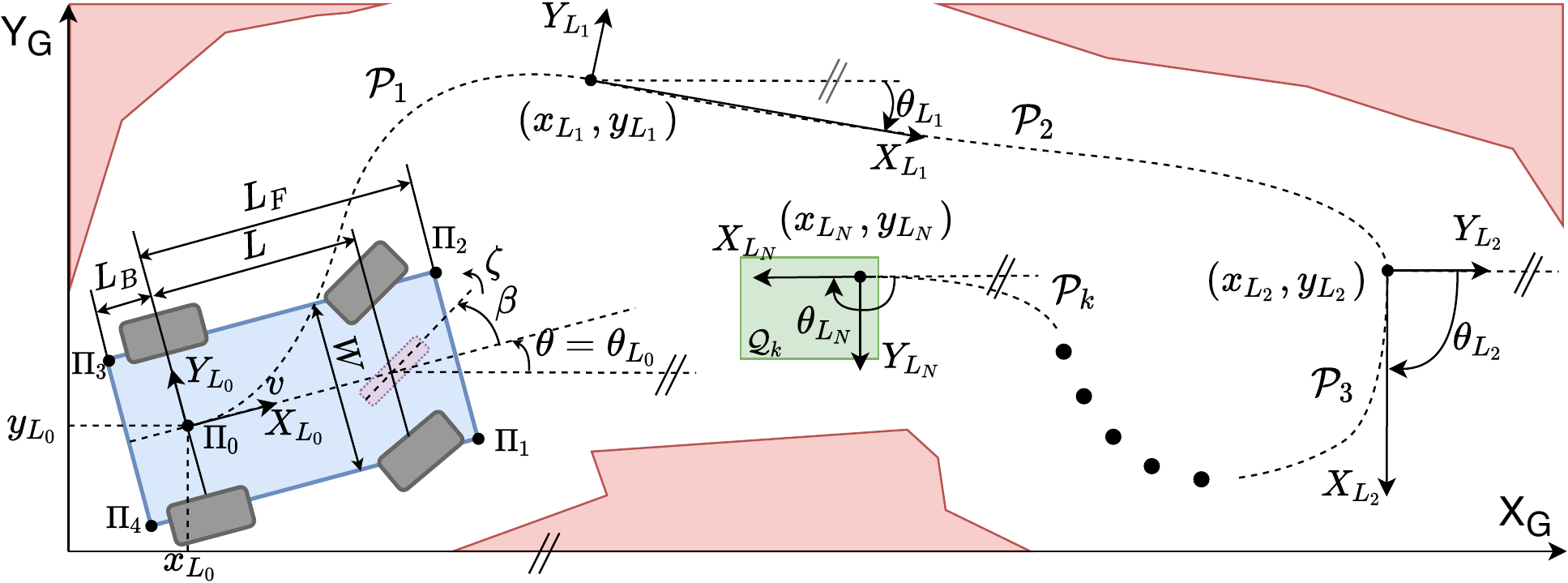}
    \caption{A scheme of the proposed problem. Planning a monotonic and feasible path 
        from an initial state ${q_0 = \begin{bmatrix}\beta_{L0} & \theta_{L0} & x_{L0} & y_{L0}\end{bmatrix}^T}$ to a subset $\mathcal{Q}_k$ of desired configurations (the green rectangle). Depicted exemplary solution is a
    spline built with the 5-th degree polynomials defined in the local coordinate systems associated with the segment's endpoints.}
    \label{fig:problem}
\end{figure*}

In this paper, we focus on path planning for a kinematic car with body modeled as a rectangle (Figure~\ref{fig:problem}) with typical kinematic constraints; no lateral and longitudinal slip, and limited steering angle $\beta$. Thus, the robot state can be defined by $q = \begin{bmatrix}\beta & \theta & x & y\end{bmatrix}^T \in \mathcal{Q}$, where $\theta$, $x$ and $y$ define vehicle orientation and position in the global coordinate frame. 
We consider a problem of planning a feasible monotonic, curvature-continuous path $\mathcal{P}$ from an initial state $q_0 \in \mathcal{F}$ to a subset $\mathcal{Q}_k \subset \mathcal{F}$ of desired configurations (which are visualized by a green rectangle in Figure~\ref{fig:problem}), where $\mathcal{F} \subset \mathcal{Q}$ is a free subspace of the configuration space $\mathcal{Q}$. 
The vehicle is controlled using an input vector $u = \begin{bmatrix}\zeta & v\end{bmatrix}^T$, where $\zeta$ is angular velocity of the virtual steering angle, and $v$ is linear velocity of the guiding point $\Pi_0$. The vehicle's kinematics can be written by
\begin{equation}
    \dot{q} = 
    \begin{bmatrix}
    \dot{\beta} \\
    \dot{\theta} \\
    \dot{x} \\
    \dot{y} \\
    \end{bmatrix} = 
    \begin{bmatrix}
    1 & 0 \\
    0 & \frac{\tan\beta}{L} \\
    0 & \cos\theta \\
    0 & \sin\theta \\
    \end{bmatrix}
    \begin{bmatrix}
    \zeta \\
    v
    \end{bmatrix}.
\end{equation}

We consider a path $\mathcal{P}$ to be feasible when it is collision-free concerning the whole vehicle body (which is depicted by a blue rectangle in Figure~\ref{fig:problem}), and it has limited curvature, so that it can be followed by a car with a limited steering angle, and ends in the subset of desirable configurations $\mathcal{Q}_k$.
In our setup, the environment $E$ is modeled with a $128\times 128$ occupancy grid with resolution of $\SI{0.2}{\metre}$, thus we consider a $\SI{25.6}{\metre} \times\SI{25.6}{\metre}$ local map.

\section{Proposed Solution}
The general scheme of the proposed solution to the problem described in Section \ref{sec:problem_formulation} is presented in Algorithm \ref{alg:solution}.
Crucial parts of this algorithm, such as representation of the path $\mathcal{P}$, construction of the policy $\pi_\phi$ as well as the loss function $\mathcal{L}$ and the data used to train the proposed system are described in the following subsections.

\begin{algorithm}[h]
\SetAlgoLined
 Given: Map $E$, initial configuration $q_0$, desired configuration $q_d$, initial policy parameters $\phi$\;
 \For{$i\gets1$ \KwTo $N$}{
     Evaluate policy $\pi_\phi(E, q_0, q_d)$ to obtain parameters $S_i$ of the $i$-th segment endpoint\;\label{alg:evaluate}
     Perform virtual nominal move along the path segment and update $q_0$ to the new vehicle configuration\;\label{alg:move}
 }
 Parameters $S$ determined by the policy $\pi_\phi$ define a path $\mathcal{P}$\;\label{alg:path}
 Calculate the loss $\mathcal{L}$ of the path $\mathcal{P}$\;\label{alg:loss}
 \If{training}{
   Calculate the gradient of $\mathcal{L}$ with respect to $\phi$\;
   Update policy parameters $\phi$\;
 }

 \caption{Our proposed path planning algorithm.}
 \label{alg:solution}
\end{algorithm}

\subsection{Path Representation}
In the proposed solution, a path $\mathcal{P}$ is represented by a spline of the 5-th degree polynomials as they have a sufficient level of smoothness, so they may be tracked by a car with a continuous steering angle.
The number of spline segments ($N$) used to form a path is a free parameter, which depends on the length and complexity of resulting plans. This parameter is a trade-off between the ability to represent complex maneuvers and the path computation time, as it scales linearly with it. 

In order to represent the resulting path as a level-curve, which is required by the control algorithm \cite{levelcurve}, we defined subsequent segments by functions of the form $y = f_i(x) = \sum_{j=0}^{5} a_{ij} x^j$ for $i \in \{0, 1, \ldots, N-1\}$, where $a_{ij}$ are the polynomial coefficients, and both $x$ and $y$ are expressed in the local coordinate frame $L_i$ associated with the $i$-th segment endpoint (see Figure~\ref{fig:problem}).
Our formulation enables us to represent paths which are not the functions of the $x$ coordinate. However, it limits the maximum possible turn range during the maneuver to $N\frac{\pi}{2}\SI{}{\radian}$, which is enough for most of the typical local car maneuvers, for sufficiently large $N$.

Assuming that we know the initial state $q_0$, we can define the aforementioned spline using a $N \times 4$ matrix $S$ of parameters of segment endpoints. Each row of the matrix defines a 2D position $(x, y)$ of an endpoint as well as the first $\frac{dy}{dx}\rvert_{(x, y)}$ and the second derivative $\frac{d^2y}{dx^2}\rvert_{(x, y)}$ of the path at that point. All parameters in the $i$-th row of the $S$ are expressed with respect to the $i-1$-th local coordinate system $L_{i-1}$.

\subsection{Policy Approximation}
\label{sec:policy approximation}
To solve the problem described in Section \ref{sec:problem_formulation}, instead of applying a search algorithm to every single problem instance, we rather reformulate the problem by a Markov Decision Process (MDP),
and we aim to find a policy $\pi_\phi(E, q, q_k)$, parameterized with parameters $\phi$, which is able to solve a large group of instances of this problem.
In the MDP formalism, the agent (a vehicle in our case) takes states as input,
and generates actions that influence immediate rewards, and subsequently, future states and rewards.
We define the MDP by specifying;
\begin{itemize}
    \item the state space ${\cal S}$ that contains all possible states of the environment $s \in {\cal S}$, 
    \item the action space ${\cal A}$ that contains all possible actions the agent can execute $a \in {\cal A}$, and 
    \item the immediate reward $r(s,a)$ received by the agent if it takes the action $a$ in the state $s$.
\end{itemize}

A typical MDP \cite{suttonbarto} is defined as a tuple $\langle{\cal S},{\cal A}, P,r\rangle$,
where $P(s_{t+1}|s_{t},a)$ represents the conditional probability of transition to the next state $s_{t+1}$ given that a specified action is taken at the current state $s_{t}$ under the action $a$. In the path planning problem, we can assume that the transitions are deterministic, as long as the planned paths are feasible.
Therefore, in our formulation, the states consist of the current $q$ and the desired (goal) $q_k$ configurations of a vehicle, and the environment (scene) map $E$.
Within a planning episode (i.e. a single maneuver) the map and the goal state remain unchanged, while the current state evolves along the planned path.
The actions are defined as path segments (see line \ref{alg:evaluate} of Algorithm \ref{alg:solution}),
transitions are deterministic moves along the planned path segments (see line \ref{alg:move} of Algorithm \ref{alg:solution}), while the rewards are dependent on the collisions, path curvature, and reaching the goal (see line \ref{alg:loss} of Algorithm \ref{alg:solution}).
Considering the deterministic transitions, we limit the scope of our work to employing a family of deterministic policies.
Hence, one can see that these policies are in fact functions, which we call \textit{planning functions}.

Let's consider a special type of planning function, which returns a feasible path for any input if such a path exists. We will call this special function an \textit{oracle planning function}.
Knowing and being able to evaluate such a function enables us to solve all planning problems that are covered by our definition.
Although it is impossible to give a closed formula for such a function in general,
the abilities of human drivers to learn how to steer effectively during rapid local maneuvers
suggest that the oracle planning function may be approximated in some range
of its arguments, namely, for short paths in a set of environments drawn from similar distributions, using the prior experience.
This approach transfers the problem of planning into the problem of approximation of the oracle planning function.

Neural network models can approximate nonlinear functions of many variables when they are trained using statistically sufficient training data \cite{hornik,Cybenko89}.
Moreover, their inference time is usually short and stable between calls.
Therefore, we propose to learn a deterministic policy $\pi_\phi(E, q, q_k)$ which is
parameterized with a neural network, and to use it as the estimate of the oracle planning function in the rapid path generation problem.
In our considerations, the action is defined by the parameters of the segment endpoint $S_i$,
and the reward is differentiable, as it is the loss function defined in Section \ref{sec:loss} taken with the minus.
Moreover, we assume that throughout an episode, both the environment representation $E$ and the desired
configurations remain fixed, and the number of steps within an episode is equal to $N$.

\begin{figure*}[th]
    \centering
    \includegraphics[width=0.9\linewidth]{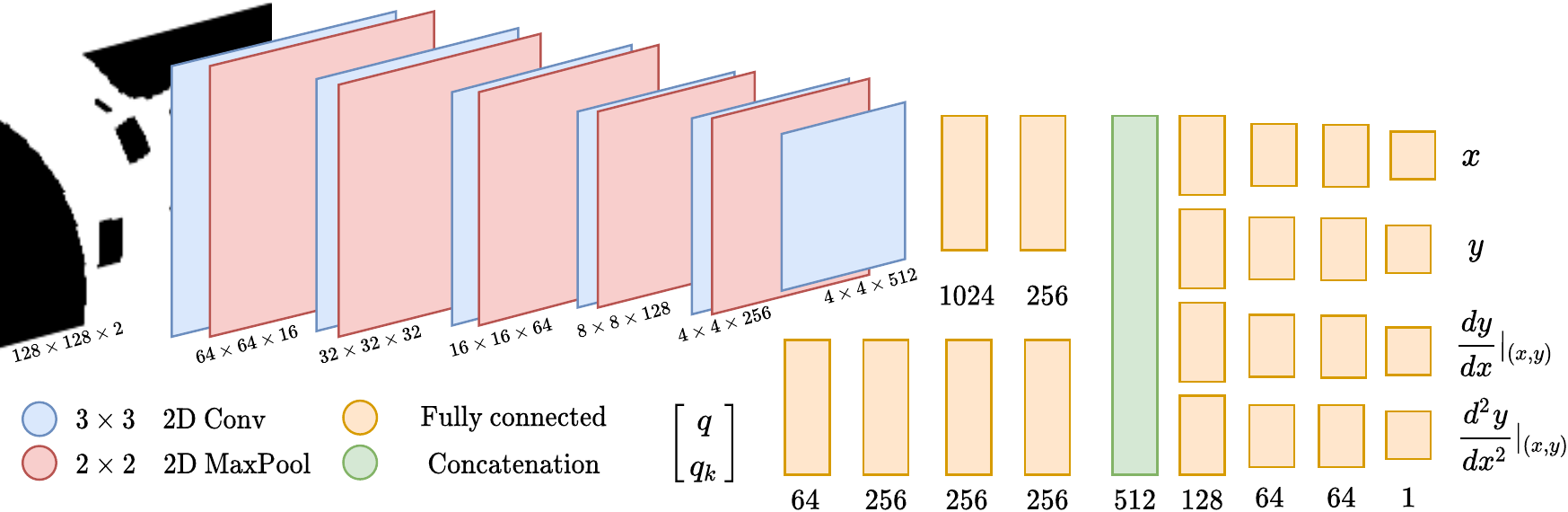}
    \caption{Architecture of the proposed neural network which is used to generate path segments.}
    \label{fig:nn}
\end{figure*}

We propose a neural network with the architecture presented in Figure~\ref{fig:nn}. It consists of 3 main parts: (i) map processor, (ii) configuration processor, and (iii) parameter estimator.  The map processor consists of a series of convolutions, max-pooling layers, and fully connected layers. The map processor takes the map as an input and produces the latent representation of the environment.
Similarly, configuration processor transforms the representations of both actual ($x, y, \sin\theta \cos\theta, \beta$) and desired ($x_k, y_k, \sin\theta_k, \cos\theta_k$) configurations using a sequence of fully connected layers. Then, latent representations are concatenated and based on this, the segment endpoint parameters are determined. Each parameter has its own \textit{network head}, and they differ only in terms of the last layer, where for $y$, $\frac{dy}{dx}\rvert_{(x, y)}$, and $\frac{d^2y}{dx^2}\rvert_{(x, y)}$ heads, there is no activation. For $x$ head, there is a sigmoid at the end, and its output is scaled $10$ times in order to set maximal length of the single path segment to \SI{10}{\meter} and stabilize gradient propagation. Except for these layers, the rest of the fully connected layers use $\tanh$ non-linearity, while convolutional layers utilize ReLU activations.
Such a network calculates the parameters of a single segment endpoint. Thus, in order to generate a spline made of $N$ polynomial segments, we perform $N$ inferences by changing the representation of the actual state $q$ each time, according to the previously determined segment.

\subsection{Our Proposed Differentiable Loss Function}
\label{sec:loss}

\begin{figure}[th]
    \centering
    \includegraphics[width=0.5\linewidth]{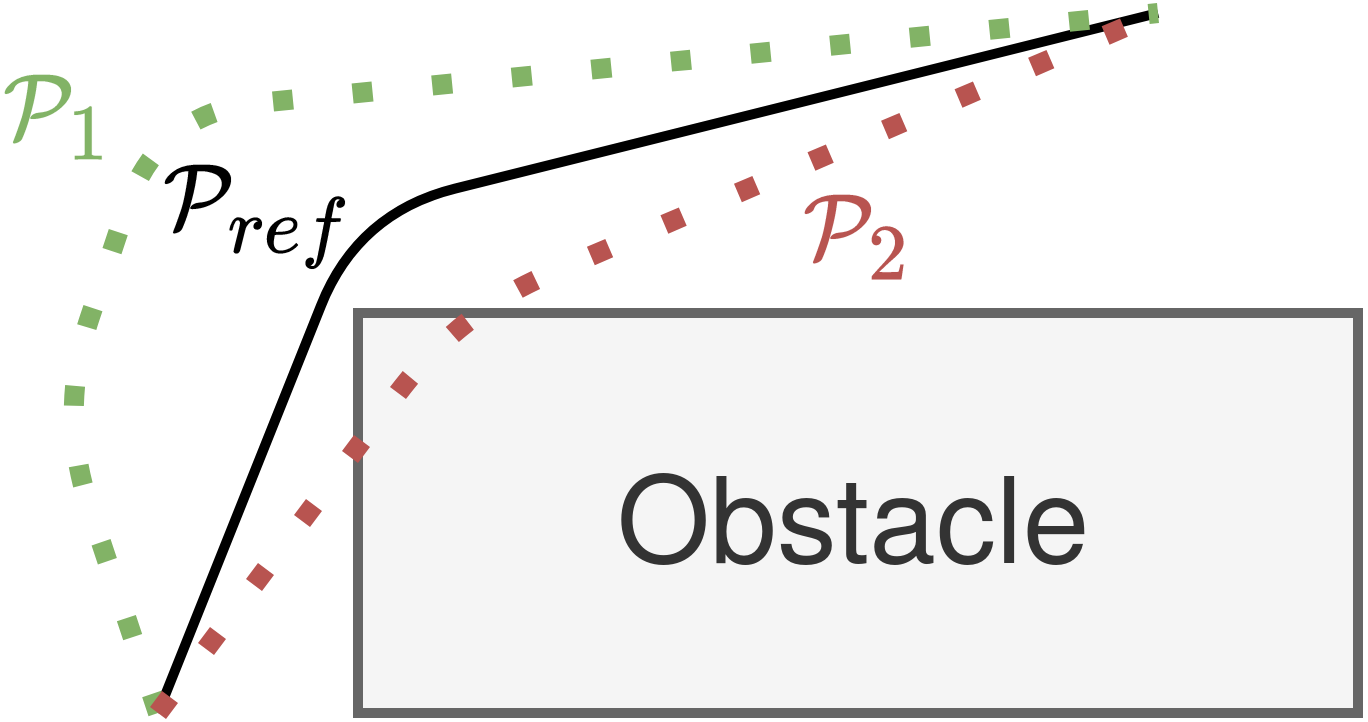}
    \caption{The path $\mathcal{P}_2$ has lower error in terms of imitation learning (lies closer and is more similar to the reference path $\mathcal{P}_{ref}$) than the path $\mathcal{P}_1$. However, the path $\mathcal{P}_2$ is infeasible, in contrast to $\mathcal{P}_1$. Such a situation can be omitted using Reinforcement Learning.}
    \label{fig:sl_error}
\end{figure}

To train the proposed neural network, we need to define a loss function, which will drive our model towards some \textit{oracle planning function}. The probably simplest solution is to use Imitation Learning to copy the behavior of a complete planning algorithm on multiple path planning tasks. However, this approach has several important drawbacks. First, the usage of a planning algorithm requires a significant amount of time to generate training data. Moreover, the performance of the chosen planner constitutes an upper bound on the proposed planning policy, which results in paths far from optimal paths, or increases the time needed to create the training data. Furthermore, the notion of error in supervised learning drives the solutions towards the planned paths without taking care of their actual feasibility, which may be an issue for models with limited capacity, which usually make some errors. Such a situation is schematically illustrated in Figure~\ref{fig:sl_error}.

Due to the aforementioned drawbacks of the imitation learning, we exploited the Reinforcement Learning paradigm in the form of Gradient-based Policy Search \cite{policy_search}. Thus, we constructed a differentiable loss function, which encourages the model to act as an \textit{oracle planning function} -- generate feasible paths, to learn the desired behavior.

The proposed loss function has four components:
\begin{itemize}
    \item Collision loss $\mathcal{L}_{coll}$, which is used to force the planner to produce collision-free paths.
    \item Curvature loss $\mathcal{L}_{curv}$, which is used to ensure that the produced paths are possible to follow by the car with a limited steering angle.
    \item Overshoot loss $\mathcal{L}_{over}$, which is used to ensure that the last segment endpoint lies within a neighborhood $\mathcal{Q}_k$ of the goal configuration $q_k$.
    \item Total curvature loss $\mathcal{L}_{tcurv}$, which is a regularization term used to ensure that the feasible paths have reasonably low path total curvature, as it allows for path following with greater velocity.
\end{itemize}
All losses are summed together to obtain the overall loss, except the total curvature loss $\mathcal{L}_{tcurv}$, which is added only if path is feasible, that is, $\mathcal{L}_{coll} + \mathcal{L}_{curv} + \mathcal{L}_{over} = 0$.

All the aforementioned loss components are rather not standard, so, we introduce their definitions.
For both collision and curvature losses, we consider a path as a sequence of $N$ segments, and we divide each segment with 128 points and with orientation, equally distant from each other along the $x$ axis of the local coordinate system.
For each of these points, five characteristic points $\Pi_{ijk}$ on the vehicle model (four corners of the rectangular body of the vehicle and the guiding point in the middle of the rear axle) are calculated \cite{distpoints} and used for calculation of the collision loss.
Index $i$ denotes the index of a segment, $j$ denotes the index of a point in the segment, whereas $k$ denotes the index of the characteristic point ($0$ denotes the guiding point, and the points numbered from $1$ to $4$ are the corners -- see Figure~\ref{fig:problem}).
Then, the collision loss is defined by
\begin{equation}
\label{eq:collision_loss}
 \mathcal{L}_{coll} = \sum_{i=1}^{N} \sum_{j=1}^{127} \sum_{k=1}^{5} \sigma(\Pi_{ij}, \mathcal{F}) d(\mathcal{P}_r, \Pi_{ijk}) l_{ij},
\end{equation}
where $\mathcal{P}_r$ denotes the reference path, $d(X, Y)$ is the smallest Euclidean distance between $X$ and $Y$, and $l_{ij}$ is the Euclidean distance between the $(j-1)$-th and the $j$-th point in the $i$-th segment. However, those distances are taken into account only if the vehicle periphery $\Pi_{ij}$ is in collision with the environment, which was denoted in \eqref{eq:collision_loss} by $\sigma(\Pi_{ij}, \mathcal{F})$.
Collision $\sigma(\Pi_{ij}, \mathcal{F})$ is determined by checking whether the points on the circumference of the vehicle lie inside obstacles. In our experiments, we sampled points from the circumference such that they lie no further than $\SI{0.2}{\metre}$ from each other. This ensures that we are checking collisions at the same resolution as the resolution of the grid map. 

Similar to calculation of collision loss, curvature loss $\mathcal{L}_{curv}$ and total curvature loss $\mathcal{L}_{tcurv}$ are also calculated using the division of segments by
\begin{equation}
  \mathcal{L}_{curv} = \sum_{i=1}^{N} \sum_{j=1}^{127} \max(|\kappa_{ij}| - \kappa_{max}, 0) l_{ij}
\end{equation}
and
\begin{equation}
\label{eq:tcurv}
 \mathcal{L}_{tcurv} = \gamma \sum_{i=1}^{N} \sum_{j=1}^{127} |\kappa_{i(j+1)} - \kappa_{ij}|,
\end{equation}
where $\kappa_{ij}$ is a curvature of the path at the $j$-th point in the $i$-th segment, $\kappa_{max}$ is maximal admissible path curvature, and $\gamma$ is a regularization parameter, which we set to $10^{-4}$.
In turn, the overshoot loss $\mathcal{L}_{over}$ is defined as a Manhattan Distance \citep{manhattan} from the subset of the desired configurations $\mathcal{Q}_k$.

All parts of the loss function are differentiable almost everywhere, which enables us to train it directly using the gradient of the proposed loss function in a Reinforcement Learning paradigm.
Among these loss components, there is only one, which cannot be calculated using the features of the generated path, thus cannot be trained using Reinforcement Learning methods exclusively, and it is the collision loss  $\mathcal{L}_{coll}$. To evaluate \eqref{eq:collision_loss}, we need a reference path $\mathcal{P}_r$, which we obtain from some other planner. However, since the supervision is used only to escape from the forbidden areas of the state space and it is not a performance upper-bound, we call it \textit{weak supervision}. Thus, we called the proposed training procedure \textit{Weakly Supervised Gradient-based Policy Search}.

\subsection{Our Proposed Motion Planning Dataset}
In order to train our model in a weakly-supervised manner, we need a dataset of local planning problems.
These problems should be solvable using a complete path planner employing our representation of the path.
Therefore, the purpose of using an auxiliary planner in our method is twofold:
Firstly, once a path is planned, we know that the particular task and scene combination is solvable,
and it is safe to include it in the training data. Secondly, we can use the
planned reference path to drive the learned solution outside the collision regions.
To generate such a dataset, we harvested maps of some real urban environments.
Using the Velodyne HDL-64E LiDAR measurements from the KITTI dataset \cite{kitti}
and data we have acquired in a town suburbs with a Sick MRS-6124 LiDAR mounted on a bus,
we generated a set of occupation grid maps.
In all cases, the LiDAR scans were registered using the LiDAR Odometry and Mapping (LOAM) algorithm \cite{loam},
and the 2D drivable terrain maps were obtained processing the registered laser scans with an elevation mapping method that handles properly sparse laser data \cite{ps2012}.
Finally, 2D occupancy grids were produced from the elevation maps by thresholding the elevation
values and setting the drivable (empty) and non-drivable (occupied) cells.
In addition, we used CARLA simulator to gather, much less noisy maps, with multiple obstacles such as pedestrians and different vehicles.
We built some maps of the Town05 and Town07 to capture both city and rural areas.
All the aforementioned maps have the same grid resolution of 0.2~m/px.

\begin{figure*}[th]
    \centering
    \includegraphics[width=0.95\linewidth]{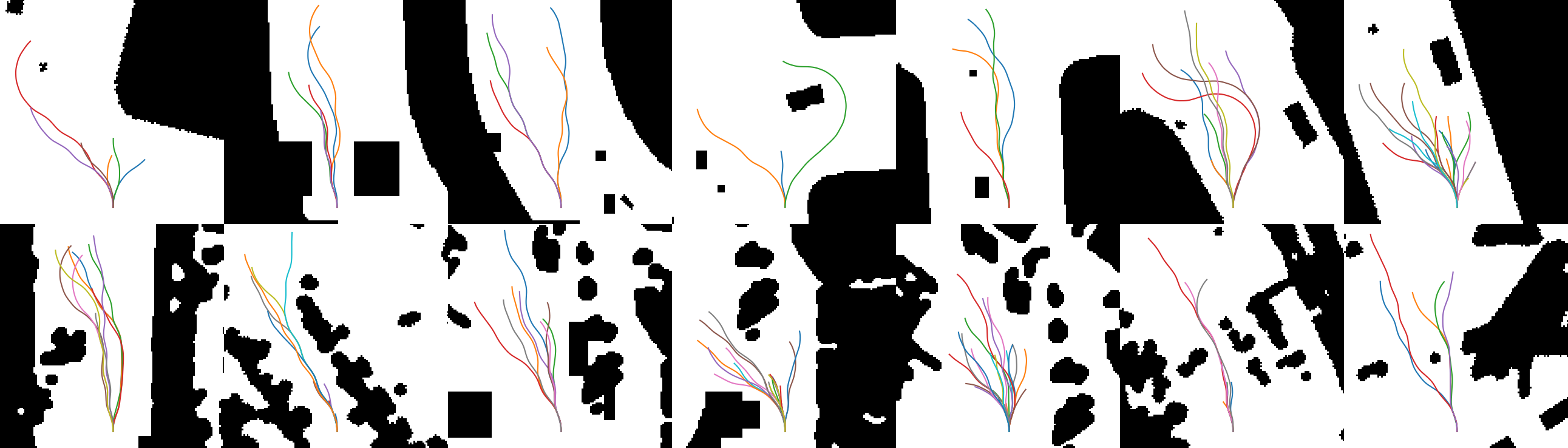}
    \caption{Visualization of sample maps of the proposed dataset. Maps together with exemplary paths planned with the State Lattices algorithm, which connects the actual robot state with some randomly drawn configurations. Scenarios in the first row are obtained from CARLA Town05 and Town07, whereas in the second row, we show maps which are obtained from real LiDAR data.}
    \label{fig:ds}
\end{figure*}

From all those maps, we sampled vehicle centered local maps (128$\times$128 px). We tied the global coordinate system in the middle column and the 120-th row of the map image, oriented upwards, and situated the vehicle there. The obtained local map could be augmented with up to 15 randomly placed rectangular obstacles. 
For each local map, we generated multiple plans with random final configurations. The plans have been obtained using a modified State Lattices (SL) planner utilizing 73 polynomial path primitives. To speed up the generation process, we guided the state lattice search using a Dubins distance as a heuristic. Such a lattice-based approach makes it easy to ensure that the time and memory needed to generate a plan are bounded. Furthermore, one can easily tune the search resolution (it corresponds to the number of primitives) and can guarantee the desired accuracy of reaching the final configuration (here 0.2 m in terms of Dubins distance). Sample problems of the proposed dataset (grid maps and auxiliary reference paths) are visualized in Figure~\ref{fig:ds}.

In this manner, we generated 23517 local maps and 134509 scenarios.
Firstly, we excluded the test set, which was generated entirely from a single map of our own Sick LiDAR dataset. It contains 1014 local maps and 8128 scenarios.
Secondly, we collected maps using; (i) another map from our dataset, (ii) maps built from the KITTI dataset, and (iii) maps generated in CARLA.
Then, we randomly selected 1996 maps to construct a validation set, and the rest of the maps was used as a training set. Finally, we generated paths for those maps, and obtained 11008 and 115319 scenarios in the validation and training set, respectively.
We examine the generalization ability of the models to previously unseen environments by using data collected from entirely different environments in the training and test sets.
We would like to share our dataset with the community as a benchmark for kinematically constrained motion planning\footnote{Dataset is available at \url{https://drive.google.com/file/d/1TACIbwO6L4qTL3lzfvcPc3HDiVmxtd77/view?usp=sharing}}.

\section{Experimental Results}
In our experiments, we modeled a Kia Rio III, whose physical dimensions are the following: distance from rear axle to rear bumper is $L_B = \SI{0.67}{\metre}$, distance from rear axle to front bumper is $L_F = \SI{3.375}{\metre}$, distance between axles is $L = \SI{2.8}{\metre}$ and vehicle width is $W = \SI{1.72}{\metre}$. Moreover, for the length $L$ and maximal steering angle $\beta_{max} = \SI{0.57}{\radian}$,  the maximal admissible path curvature is $\kappa_{max} = \SI{0.227}{\per\metre}$. The subset of the desired states $\mathcal{Q}_k$ is defined around a given desired state $q_k$, such that its elements should not lie further than $\SI{0.2}{\metre}$ from the $q_k$ both in $x$ and $y$ axes, and the orientation $\theta$ should satisfy $\abs{\theta - \theta_k} < \SI{0.05}{\radian}$. We trained our planner using Adam \cite{adam} optimizer with learning rate equal to $10^{-4}$, and batch size of 128. The models were trained for 400 epochs. It took 33 hours to train the models using NVIDIA GTX1080Ti GPU.

\subsection{Impact of the number of segments of the generated path}
\begin{figure}[th]
    \centering
    \includegraphics[width=0.95\linewidth]{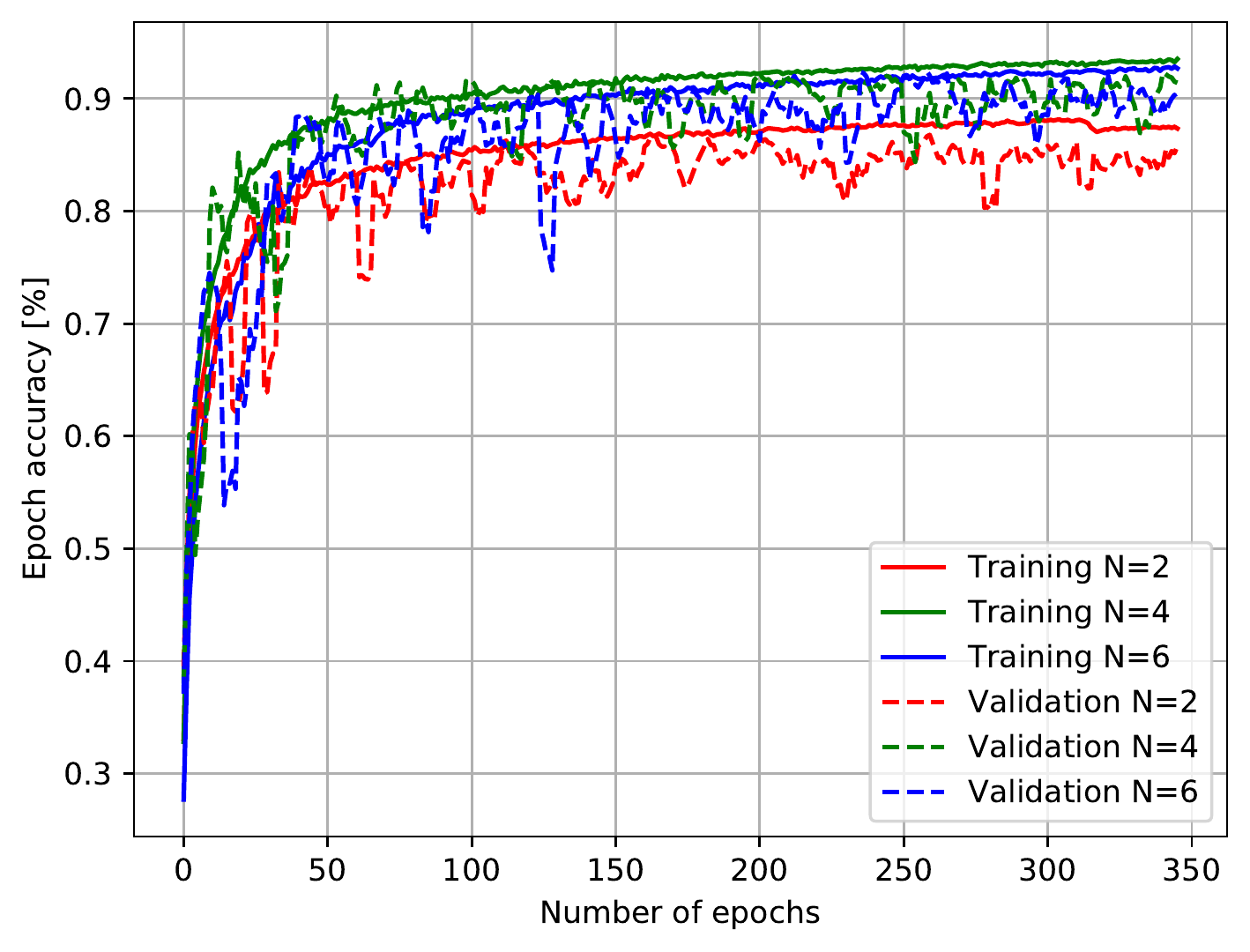}
    \caption{Change of accuracy of the models (on training and validation sets) for $N \in \{2, 4, 6\}$ through the learning process.}
    \label{fig:learning_curves}
\end{figure}

At the first stage of the experiments, we evaluated how the number of path segments $N$ affects the learning process, runtime, and accuracy of the proposed machine learning based path planner. Figure~\ref{fig:learning_curves} presents the learning curves of the models (i.e., change of their accuracy on both training and validation sets)  trained with a parameter $N \in \{2, 4, 6\}$. One can see that the accuracy of all models grows rapidly from the very first epoch, and achieves 80\% in less than 40 epochs. The fastest learning rate was obtained for $N=4$, however, the learning curves approach a similar level of performance for both $N=4$ and $N=6$ at the end. Notice that the accuracy of models obtained for the validation sets is only slightly worse than for the training sets, and it does not decrease over time. This observation may suggest that the models have learned some transformation from the task definition into the solution path, which is applicable beyond the training examples to unseen desired poses and environment maps.

\begin{figure}[th]
    \centering
    \includegraphics[width=0.75\linewidth]{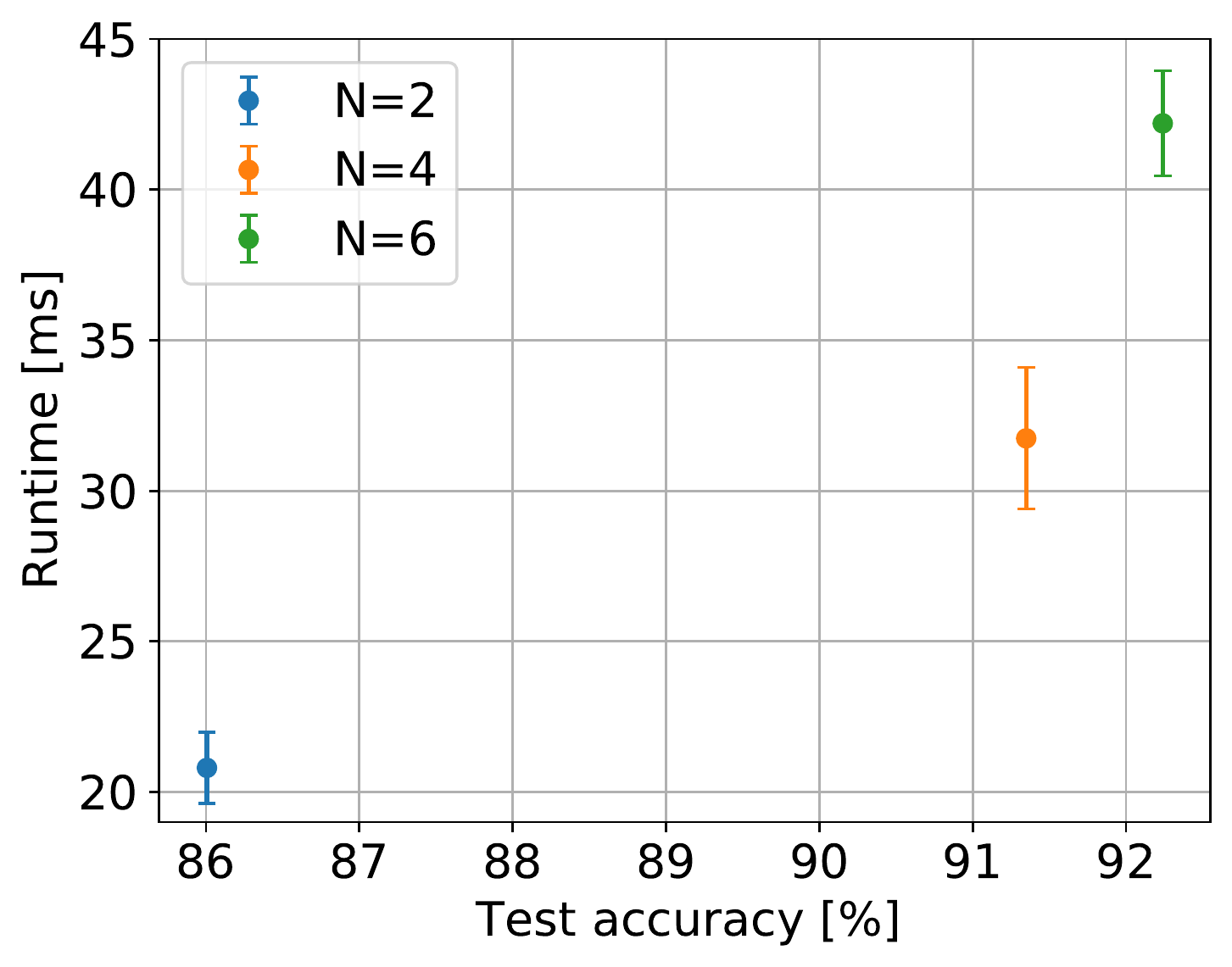}
    \caption{Mean inference time and their standard deviation, as well as the test set accuracy of the proposed neural networks for $N \in \{2, 4, 6\}$.}
    \label{fig:time_vs_acc}
\end{figure}

The number of the predicted segments $N$ affects also the inference time. In Figure~\ref{fig:time_vs_acc}, we present the mean and standard deviations of model running times, for different $N$ values, together with its accuracy on the test set. One can see that models trained with larger $N$ are slower, but achieve superior performance. Therefore, we observe a tradeoff between the performance on the test set and the inference time.
The accuracy increases with larger $N$, since larger $N$ allows performing much more complicated maneuvers, which may be necessary for highly confined spaces.
Since all the obtained times are relatively low (all networks guarantee that the path computation time is below $\SI{50}{\milli\second}$), we decided to choose the neural network model trained for $N=6$ in the other experiments, as it achieved the best accuracy in the analyses, and it allows planners to perform more complex maneuvers compared to the other values.

\subsection{Validation on the test set}
\begin{figure*}[th]
    \centering
    \includegraphics[width=0.95\linewidth]{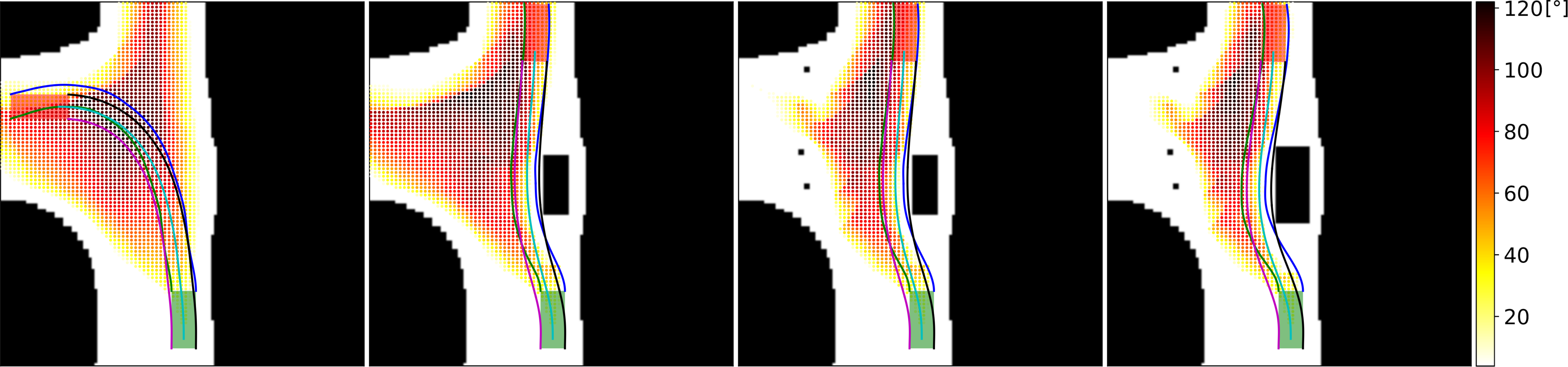}
    \caption{Visualizations of the sets of the final states (denoted by heatmaps) for which the planner generated feasible paths (so-called reachable sets). Darker dots represent wider ranges of possible end orientations of the vehicle. Colored lines depict paths drawn by the four corners of the vehicle and the guiding point while moving along the path provided by the network. The planner takes into account the obstacles located on the path to the goal (2nd image), whereas ignores the distant obstacles (3rd image). Moreover, it reacts to the obstacle size (4th image) and plans accordingly.}
    \label{fig:seq}
\end{figure*}

\begin{figure*}[th]
    \centering
    \includegraphics[width=\linewidth]{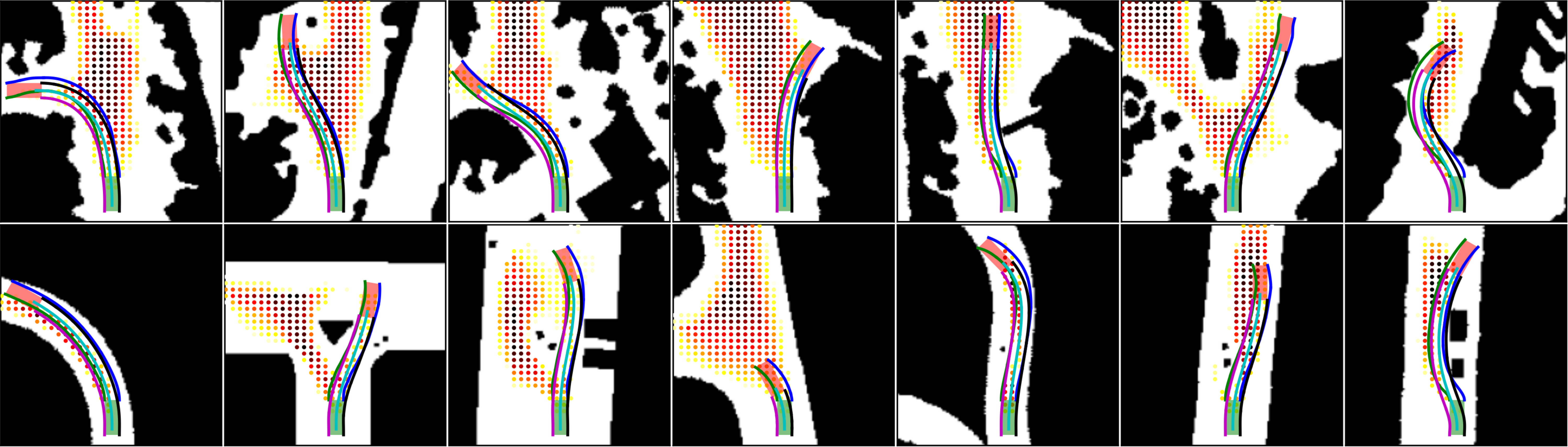}
    \caption{Visualizations of the reachable sets and exemplary paths. Maps in the first row are obtained from the test set made with real LiDAR data, whereas in the second row, we show maps which are obtained from CARLA Town04 (not used in the training).}
    \label{fig:examples}
\end{figure*}

Exemplary paths generated by the proposed planner are presented both in Figure~\ref{fig:seq} and Figure~\ref{fig:examples}, together with the sets of reachable configurations. One can see that our planner learned to plan for wide areas of the accessible state space in the previously unseen environments. Obviously, our planner is not complete. Thus, some regions of the state space are not reachable using plans generated by our network. However, it still can perform complex maneuvers in narrow passages.
Moreover, it anticipates the environment using information from the occupancy map, and it is able to detect cars, pedestrians, and other obstacles, taking their presence into account while planning (Figure~\ref{fig:seq}). However, it ignores distant obstacles by paying attention to the objects that possibly collide with the path.
We also show that our planner can use both high-quality maps obtained from the CARLA simulator, as well as lower quality maps, which are obtained from real LiDAR measurements (Figure~\ref{fig:examples}).

In Table~\ref{tab:results}, we compare the accuracy and quality of our planner to selected planning algorithms. We use the OMPL \cite{ompl} implementations of those, equipped with the Dubins curves as primitives, which makes the planning problem way easier because in this case, the graph of possible solutions is significantly smaller. Notice that our solution ensures the continuous virtual steering wheel angle throughout the path, while the rest of the planners allows for instant changes of this angle. 
In Dubins paths, curvatures change instantly from 0 to a constant value between line segments and circle arcs. Hence, Dubins paths can be followed by a car-like vehicle defined in Section \ref{sec:problem_formulation}
only if we apply an infinite steering input or stop the vehicle to change the steering angle.
Therefore, in our case, we cannot utilize a convenient steering function, which can connect two arbitrary vehicle configurations with Dubins curves. When such a steering function is not present, the performance of sampling-based planners decays rapidly \cite{NonholonomicRRT}. 

For accuracy, we compute a ratio of valid plans to all scenarios from the test set. We consider that the plan is invalid, if it violates kinematics or environmental constraints, or if the planner exceeds the maximal allowed planning time, which is set to $\SI{50}{\milli\second}$.
The proposed method obtained the best accuracy among other planners. However, it produces paths, which are on average a bit longer and require more turns. It is known that Dubins curves optimize the path length. However, its convergence is not so clear in terms of accumulated turns. We conjecture that our solution results in more turns, because of insufficient regularization of the total curvature, which enables planning slightly wavy paths, even if there is no need for these micro-turns.

\begin{table*}[!th]
\centering
\caption{Accuracy, accumulated turn and length of the paths generated by different planning methods for the $\SI{50}{\milli\second}$ planning time limit. Both accumulated turns and lengths are reported only for paths valid for all planners (except SST and RRT* due to their low accuracy).}
\begin{tabular}{c|cccccc}
\hline
Planner & SST\cite{SST} & RRT*\cite{karaman} & AIT*\cite{ait2020} & BIT*\cite{BIT} & ABIT*\cite{abit2020} & ours\\
\hline
Accuracy [\%] & 3.95 & 15.97 & 84.12 & 84.19 & 84.15 & \textbf{92.24}\\
Accumulated turn [$\SI{}{\radian}$] & - & - & \textbf{0.63$\pm$0.43} & \textbf{0.63$\pm$0.43} & \textbf{0.63$\pm$0.43} & 0.78$\pm$0.55 \\
Length [$\SI{}{\meter}$] & - & - & \textbf{11.6$\pm$6.1} & \textbf{11.6$\pm$6.1} & \textbf{11.6$\pm$6.1} & 11.8$\pm$6.2 \\
\hline
\end{tabular}
\label{tab:results}
\end{table*}

\begin{figure}[th]
  \centering
  \includegraphics[width=0.95\linewidth]{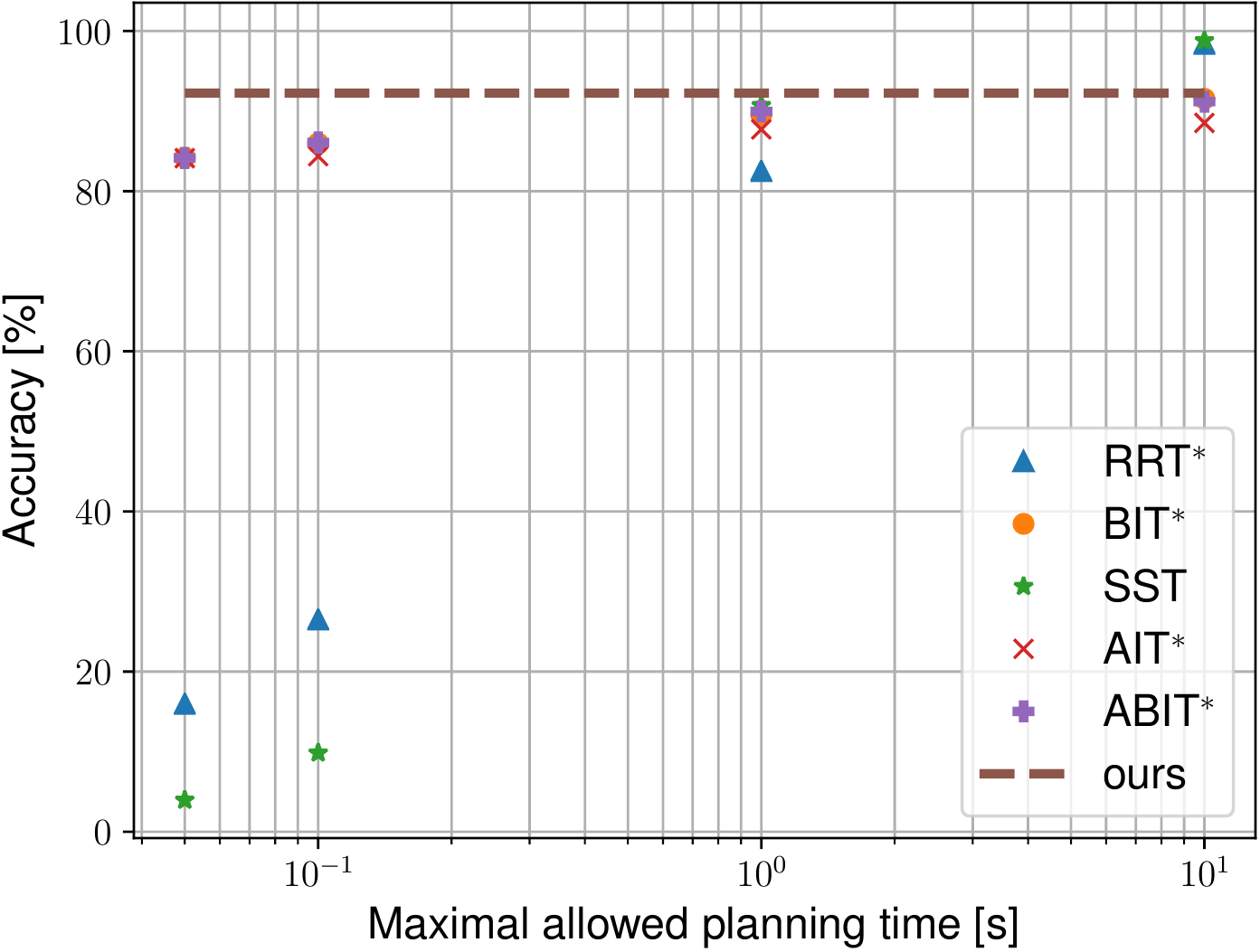}
  \caption{Accuracy [\%] for all planners on the test set with respect to the maximal allowed planning time~[$\SI{}{\second}$]. Our planner outperforms all other planners when the planning time horizon is below $\SI{1}{\second}$.}
  \label{fig:acc}
\end{figure}

A more detailed comparison of the accuracy of the planners is presented in Figure~\ref{fig:acc}. This chart gives an interpretation of how the accuracy of the given planner changes with respect to the maximal allowed planning time.
The proposed planner outperforms all other tested planners for the planning horizon up to $\SI{1}{\second}$. But, even for the $\SI{10}{\second}$ limit, it is more accurate than BIT*, ABIT*, and AIT*. This shows that our method solves correctly a large part of the scenarios within $\SI{50}{\milli\second}$, and its planning time is almost constant and equal to $41 \pm \SI{1}{\milli\second}$. However, our method is not able to improve after that point (as it is not meant to be able to). One can consider coupling our method together with RRT* or SST, as if there is more than $\SI{1}{\second}$ available, then they may improve the generated solution. For all these measurements, we used Intel Core i7-9750H based PC.

\subsection{Ablation study}
\begin{figure*}[th]
    \centering
    \includegraphics[width=\linewidth]{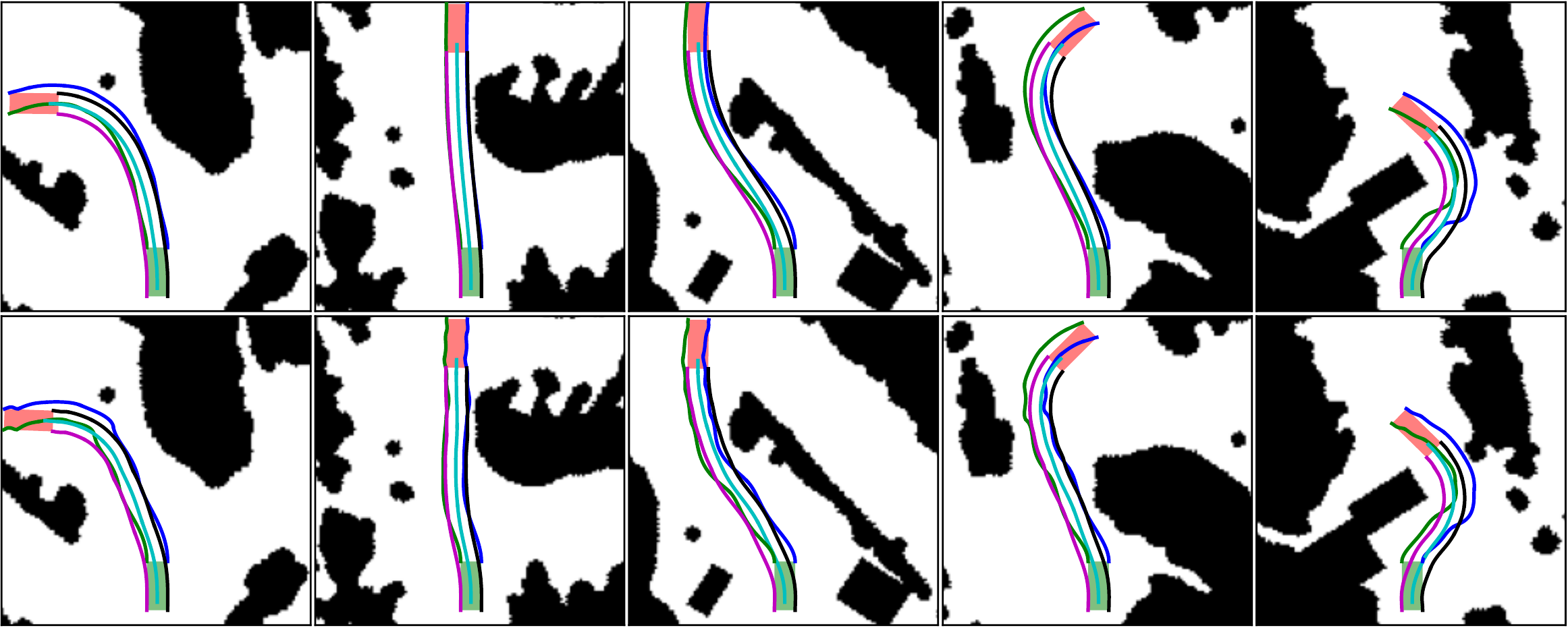}
    \caption{Exemplary paths generated by the proposed neural network based motion planner trained with the standard loss function (defined in Section \ref{sec:loss}) -- first row, and with modified loss function without $\mathcal{L}_{tcurv}$ term (defined in \eqref{eq:tcurv}) -- second row. Presence of the total curvature loss term results in much smoother looking and easier to follow paths.}
    \label{fig:ablation}
\end{figure*}

The total curvature loss $\mathcal{L}_{tcurv}$, defined in \eqref{eq:tcurv} is the only term in the loss function, which is not necessary to obtain a feasible path. Hence, we include a brief ablation study to analyze the impact of this part of our loss function (\ref{sec:loss}) on the planner accuracy and the quality of generated paths. 
For this study, we trained and tested a new model in the same conditions as the basic one (for $N=6$) but without $\mathcal{L}_{tcurv}$ in the loss function. It obtained slightly better performance in terms of the accuracy on the test set achieving 92.53\%. However, the main idea behind this loss term is visible in the comparison of the mean accumulated turn, which is equal to $\SI{0.86}{\radian}$ for the model trained without $\mathcal{L}_{tcurv}$ term, and $\SI{0.78}{\radian}$ for the model trained with $\mathcal{L}_{tcurv}$. A more intuitive view on the importance of the total curvature loss is presented in Figure~\ref{fig:ablation}, where we compare paths generated by the aforementioned models in several different scenarios. Even though in all cases the maximal admissible curvature is not exceeded, paths in the first row (generated by the model with full loss function) are even visually much smoother than those in the second row (generated by the model trained without $\mathcal{L}_{tcurv}$ term). Such behavior is desirable as it limits the sudden and frequent turns which are usually unnecessary to perform the maneuver, may limit maximal allowed velocity, and cause problems in control. Moreover, maneuvers performed smoothly, without redundant turns, are far closer to the expectations of the human drivers and give a feeling of safe driving.

\subsection{Experiments in CARLA}
Since the planning time of our method is so short, it can be used even to handle dynamic environments. We validated that claim using the CARLA simulator \cite{carla}. 
In the performed simulations, we use CARLA API to obtain the maps of the ego-vehicle local environment and the actual configuration of the ego-vehicle. Then, using that information, and a predefined list of control points, which represent the information obtained from a higher-level motion planner\footnote{In the CARLA experiments, the routes were defined manually, but in principle, any complete path planner, e.g. based on Probabilistic Road Maps can be applied here.}, the proposed maneuver generation algorithm predicts the path leading to the next control point. Finally, this path is followed using the VFO kinematic controller defined in \cite{VFO}. In such a scenario, the planner runs every $\SI{50}{\milli\second}$ and produces new paths to follow, which enables it to avoid collisions and fix possible path infeasibility resulting both from the controller errors, and inaccuracy of the network.

\begin{figure}[th]
  \centering
  \includegraphics[width=0.9\linewidth]{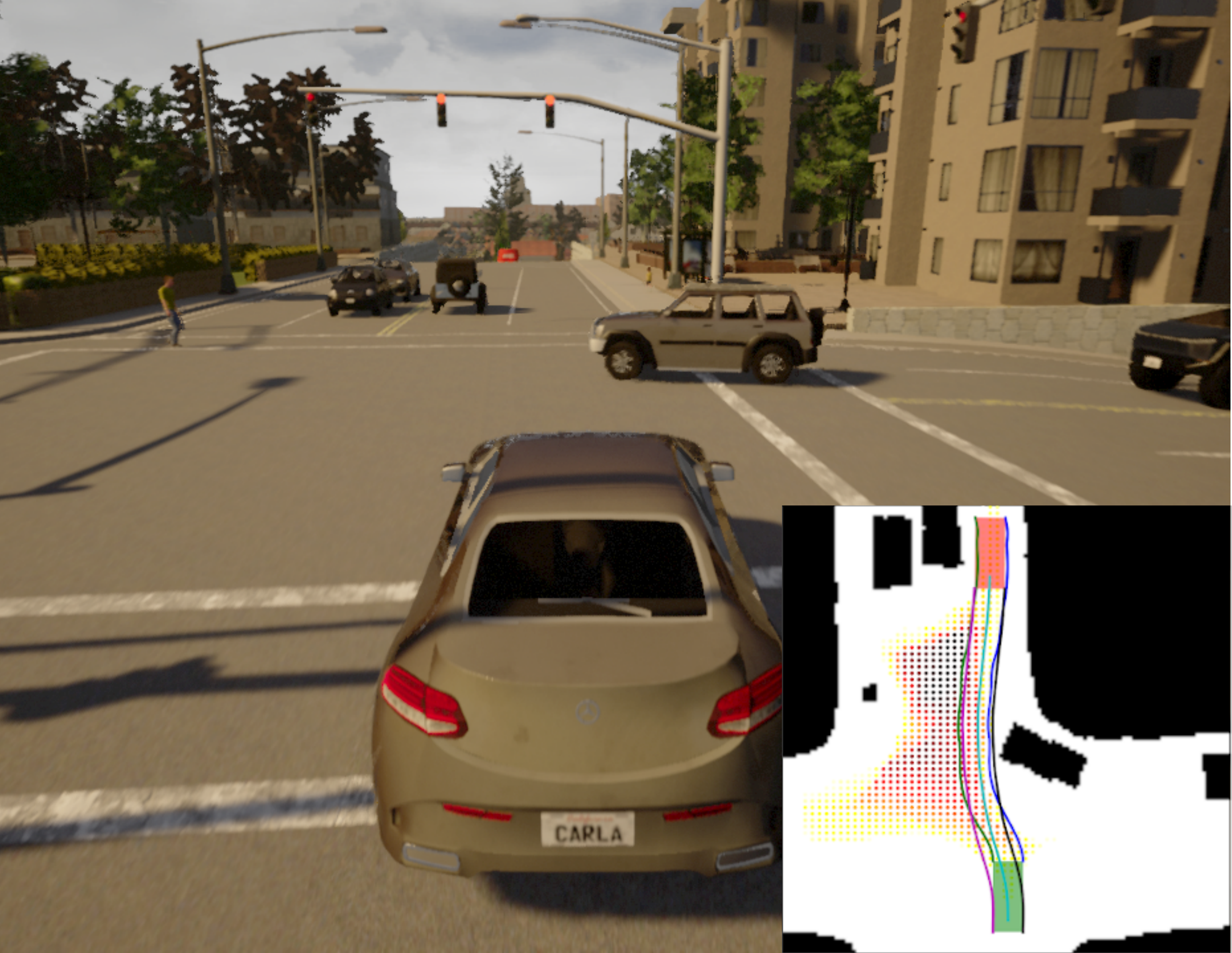}
  \caption{Visualization of the online path planning in the CARLA simulator.}
  \label{fig:carla}
\end{figure}

Figure~\ref{fig:carla} presents a view from the (simulated) camera, as well as the map given by the environment. On the map, we presented one of the possible paths to follow, together with the visualization of the accessible set.

\begin{figure*}[th]
  \centering
  \includegraphics[width=\linewidth]{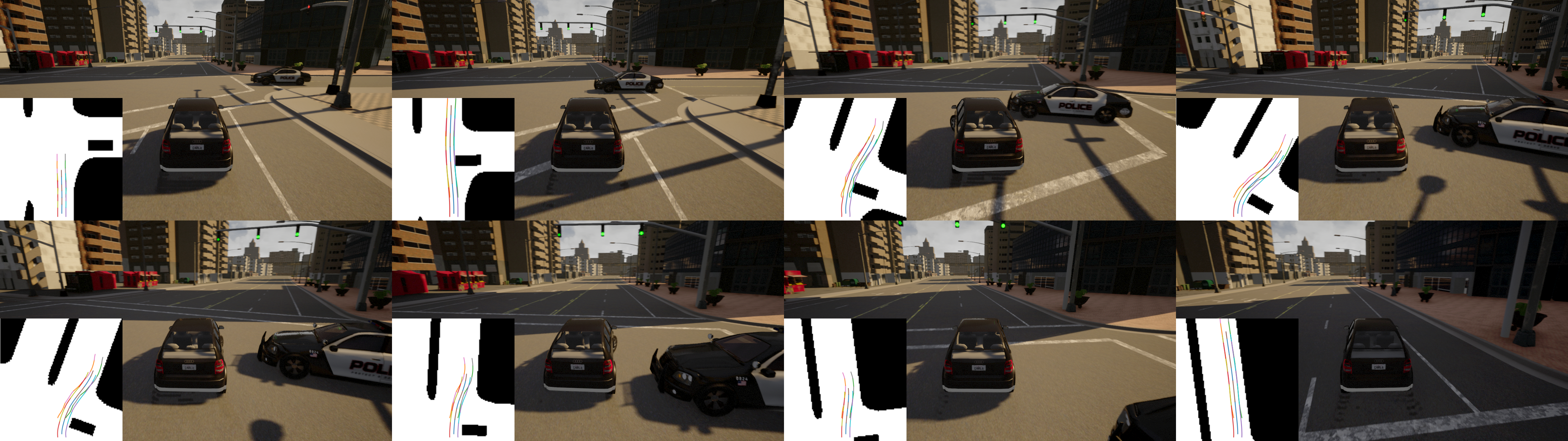}
  \caption{Visualization of the online replanning which enables the car to avoid collisions with dynamic obstacles. In the bottom left corner, there are maps, which present the planned path. This sequence is also available as a supplementary video clip (\url{https://youtu.be/dnfIp5A-k6Q}).}
  \label{fig:carla_gta}
\end{figure*}

\begin{figure*}[th]
  \centering
  \includegraphics[width=\linewidth]{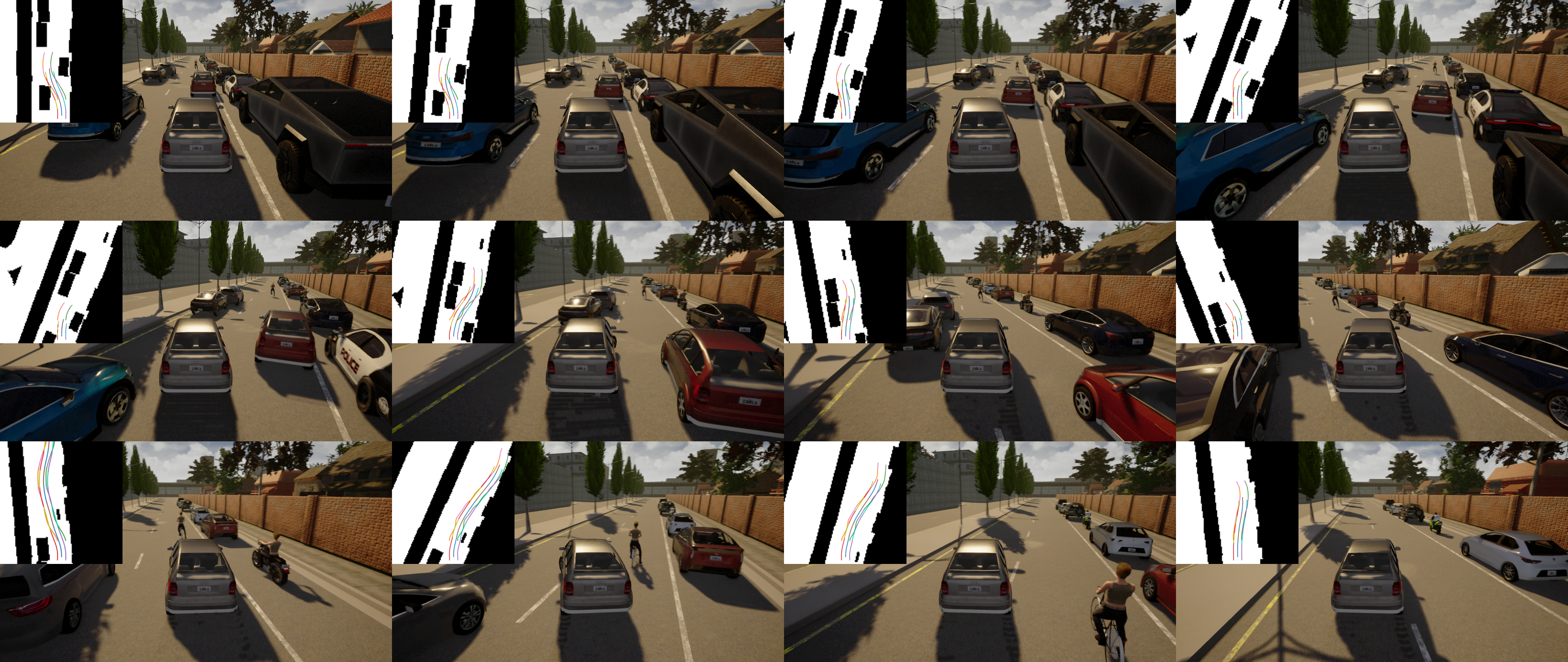}
  \caption{Visualization of the online replanning process, which enables the car to avoid collisions with static obstacles. Even though at the very first moment, the planner produces a path that collides slightly with the obstacle, after a small movement, it replans, and the produced path is feasible. In the upper left corner, there are maps, which present the planned path. This sequence is also available as a supplementary video clip (\url{https://youtu.be/pj02xUgyCrU}).}
  \label{fig:carla_traffic}
\end{figure*}

In Figure~\ref{fig:carla_gta}, we present a sequence of snapshots from the CARLA simulator taken in a scenario of crossing the intersection. 
While our car drives straight through the junction, a police car tries to block its way. The vehicle re-plans every \SI{50}{\milli\second}, thus reacts immediately (note the limited sensing range -- about \SI{25}{\metre}) and starts to follow a new path to avoid a collision.
In Figure~\ref{fig:carla_traffic}, we present a similar sequence, but for the scenario of navigation through a traffic jam. This scenario emphasizes the ability of the proposed planning algorithm to navigate in narrow passages.
Although the initial path prediction is slightly inaccurate, after a small movement along the planned path, the online re-planning process allows generating a valid maneuver.

\section{Conclusion}
We present a novel learning-based approach to local maneuver generation for self-driving cars.
Our method contests the typical approach to motion planning based on state-space search algorithms and proposes to approximate an \textit{oracle planning function} instead. Our proposed approach lets one omit the time-consuming execution of the algorithm, by learning how to plan \textit{off-line} (optimizing the approximator of the \textit{oracle planning function}), and then only reusing the gathered knowledge \textit{on-line} (using a fast approximator). Although this approach gives no guarantees about the completeness, it is fast and flexible, as its quality depends on the data used in the training process, and the expressiveness and generalization abilities of the model.
To achieve rapid path generation, an oracle planning function is approximated by a deep neural network trained using a weakly supervised gradient-based policy search.
In our approach, we do not model the \textit{oracle planning function} with the use of human-generated data, but instead, we define it implicitly by the employment of a novel loss function.

Our neural planner can handle typical local maneuvers, such as overtaking, avoiding obstacles, or parking in almost constant time, rapidly generating feasible paths that are nearly optimal with respect to their length, and ensure continuous steering angle.
The proposed approach outperforms several conventional path planners (including those very recent)
concerning the number of successfully completed tasks (achieving the accuracy of 92\%), being also by far the fastest one.
Extensive tests on local urban environments maps obtained from real LiDAR data and generated
from the CARLA simulator demonstrate good generalization abilities of the proposed network model,
and the efficiency of its weakly supervised training procedure. Experiments performed in the CARLA simulator show not only the ability of the network to produce plans, which are possible to follow with the kinematic controller but also the advantage gained because of the fast path replanning.
Ablation study presents the impact of using total curvature loss term \eqref{eq:tcurv} on the generated path smoothness.
We also provide an open-source code of our system and a public dataset of motion planning
problems with solutions that can be used by others to test the proposed planner or to compare against it.
Planned future work concerns the possibility to use our path generator combined with higher-level planners, as it is fast and accurate but not complete.



\printcredits

\bibliographystyle{cas-model2-names}

\bibliography{cas-refs}





\end{document}